\documentclass[10pt,letterpaper]{article}
\usepackage{cogsci}
\cogscifinalcopy % Uncomment this line for the final submission 

\usepackage{pslatex}
\usepackage{apacite}
\usepackage{float} % Roger Levy added this and changed figure/table
\usepackage{amsmath}
\usepackage{amsfonts}
\usepackage{amssymb}
\usepackage{graphicx}
\usepackage{subcaption}

\title{Computational Approaches to Access Probabilistic Population Codes\\ for Higher Cognition an Decision-Making}
 
\author{{\large \bf Kevin Jasberg (kevin.jasberg@uni-duesseldorf.de)} \\
  Department of Computational Linguistics, Heinrich-Heine-University \\
  Duesseldorf, 40225 Germany
  \AND {\large \bf Sergej Sizov (sizov@hhu.de)} \\
  Web Science Research Group, Heinrich-Heine-University \\
  Duesseldorf, 40225 Germany}

\begin{document}

\maketitle

\begin{abstract}
In recent years, research unveiled more and more evidence for the so-called Bayesian Brain Paradigm, i.e. the human brain is interpreted as a probabilistic inference machine and Bayesian modelling approaches are hence used successfully. One of the many theories is that of Probabilistic Population Codes (PPC). 
Although this model has so far only been considered as meaningful and useful for sensory perception as well as motor control, it has always been suggested that this mechanism also underlies higher cognition and decision-making. However, the adequacy of PPC for this regard cannot be confirmed by means of neurological standard measurement procedures.

In this article we combine the parallel research branches of recommender systems and predictive data mining with theoretical neuroscience. The nexus of both fields is given by behavioural variability and resulting internal distributions. We adopt latest experimental settings and measurement approaches from predictive data mining to obtain these internal distributions, to inform the theoretical PPC approach and to deduce medical correlates which can indeed be measured in vivo. This is a strong hint for the applicability of the PPC approach and the Bayesian Brain Paradigm for higher cognition and human decision-making.

\end{abstract}

\section{Introduction}

A modern perspective on understanding the human brain is represented by the Bayesian Brain Paradigm which states that the true origin of neural activation can never be known beyond doubt but can be inferred using theories of probability and statistics. Initially confined to  theoretical discourse, this paradigm more and more propagated into other branches of neuroscience research. The essence of this paradigm is that the brain somehow generates and processes internal probability distributions to create its very own model of the outside world. Around this assumption, some theories have emerged so far dealing with the representation of these distributions, the integration of various information channels (cue integration) and learning procedures. The most promising and unifying concept from all these directions is that of the Probabilistic Population Codes (PPC) which is in accordance with the common sense that the brain is organised in   agency. The assumption of population activity encoding probability densities has so far been positively evaluated for visual processing, auditory cue integration and motor control. It is thus conceivable that the same underlying mechanism also takes place for higher cognition and decision-making, but although many authors have postulated the PPC's involvement, no further investigations have so far been carried out to substantiate this assumption with evidence.

In parallel, recent efforts in the field of predictive data mining (e.g. recommender systems, user personalisation, individual advertising, etc.) have been made to observe, learn and predict user behaviour. Over the last few years, however, web engineers have realised that users are not constant in their interactions and denoted this behavioural variability and the associated lack of reliability of user observations as user noise or human uncertainty. Web engineers thus assumed the existence of underlying distributions, so-called feeling curves. The crux of this variability is that it becomes quite hard to tell whether a deviance between a system's prediction to a real user response is still within the limits of natural behaviour variability or not. 
More importantly, such behavioural variability is usually not covered by systems. Hence, the evaluation of research progress and particularly the detection of enhancements has become subject to some significant doubt. To account for these insights, measurement approaches have been developed to access assumed underlying distributions. These approaches together with recent suggestions for user experiments are most likely to result into fruitful data about variable decision-making that can be used for neuroscience investigations as well.

\subsection{Objectives and Research Questions}
The commonality of both research fields is to reason behavioural variability by assuming underlying probability densities. However, both fields have produced their own methods and findings, which are aggregated in this contribution to produce synergy effects.
We combine the explanatory model of PPC from neuroscience with data obtained by measurement approaches from predictive data mining as well as its methods for computer simulation. In doing so, neuroscience is gaining insight into PPC for higher cognition and decision-making. This also forms the main objective of this contribution and leads to the according research questions:
\begin{enumerate}
\item How does the multi-faceted PPC model need to be configured to be expedient for describing higher cognitions?
\item Is the configured PPC model still compatible with neurological and medical findings and hypotheses?
\end{enumerate}

\section{Related Work}

\subsection{Probabilistic Population Codes}
The basic concept relies on the so-called tuning curve, i.e. the functional relationship between a stimulus and the corresponding neuronal response. Such tuning curves have been measured to maximise for a specific stimulus, e.g. for the rotation angle of a black bar \cite{affe}. When measuring the tuning curves of multiple neurons of a rat's brain during the mammal's movement through a cage, it was found that each tuning curve maximises for another location \cite{placecells}. The discovery of these place cells suggest that neurons can be organised in such a way that their individual tuning curves fully cover the whole range of possibilities for a stimulus, so that the true state of the world can be inferred. However, one and the same stimulus never results into the same spiking behaviour of a neuron. This randomness is a conglomerate of sensory noise, cellular noise, electrical noise, synaptic noise, and noise build-up in neural networks \cite{nervousnoise}. Moreover and even more important, the authors find that ``behavioural variability, as observed in sensory estimation and movement tasks, appears to be mainly produced by noise'' \cite{nervousnoise}. In other words, human decisions can be seen as uncertain quantities by nature. All of these concepts are used in the PPC approach to explain the emergence of internal densities and their involvement in important cognitive tasks \cite{Pouget}. This model has proven itself for cases of auditory and visual cue integration \cite{cueintegration} as well as motor control \cite{motor}. To aggregate population activity and interpret internal densities, various so-called decoder functions have been proposed recently \cite{decoders} and it has been stated that it remains unclear which decoder function is veritably utilised for cognition tasks.

From this preliminary work, we assume the bell-shaped tuning curves and the peculiarity that they cover a range of possible stimuli. We also adopt the postulated decoder functions.
Contrary to previous modelling, we define variable parameters for the tuning curves and thus obtain manifold configurations for this model. In addition we question, for the first time, the reliability of those inner densities. This results into an analysis of the variability of decision-making.

\subsection{Predictive Data Mining}
One goal of predictive data mining is the prediction of human behaviour. A lot of research produced a variety of techniques and approaches which are concisely recapitulated in \cite{Jannach} and \cite{Handbook}. During the last decade, the growth of interactions continuously supported innovations in a data-driven fashion, based on user interactions and user feedback. The first reported discovery of reliability issues for this user feedback was done by \cite{Hill} through an experiment of repeated evaluations of films. 
Research on this issue recently intensified with regard to the credibility of prediction accuracy and system quality. For this purpose, frequentist and Bayesian methods of measuring user noise were developed and positively evaluated in the context of product ratings \cite{JasUMAP}. 

We copy the measurement methods and the exemplary scenario of user ratings to access the variability of decision-making. 
We also use the proposed computational methods of stochastic simulation and uncertainty propagation to generate, compare and evaluate inner distributions.

\section{Modelling Uncertain Decision-Making}
In this section we introduce the formal approach of PPC. We start by modelling a single tuning curve together with Poisson-like noise. This will be extended to a population whose noisy activity will be aggregated by decoder functions to receive a single estimation in the light of product ratings.

\subsection*{The Single Neuron Model}
The functional relationship between responses $r$ of a neuron and the characteristics $s\in S\subset\mathbb{R}$ of a stimulus is given by the so-called tuning curve $r=f(s)$. Besides irregular shapes, tuning curves have frequently been measured to be bell-shaped or sigmoid-shaped respectively. Each bell-shaped tuning curve maximises for a preferred stimulus $p:=\operatorname{argmax} f$. In our case $f\colon S \to \mathbb{R}$ can be modelled as
\begin{equation}
f_{p}(s) :=g\cdot h(p,w)(s) + o,
\end{equation}
where the bell-shape emerges from the Gaussian density function $h$ with mean $p$ (preferred value) and standard deviation $w\in\mathbb{R}^{>0}$ (tuning curve width). The additional components $g\in\mathbb{R}^{>0}$ and $o\in\mathbb{R}^{>0}$ represent a stretching factor (gain) and a spiking offset respectively.
When measuring tuning curves in reality, one will find that these are noisy and that even one and the same stimulus never leads to the same response. Neuronal responses must therefore be seen as random variables $R$. It has been found that $R\!\sim\!\operatorname{Poi}(\lambda)$ follows a Poisson distribution with expectation $\lambda=f(s)$ \cite{affe}.

\subsection{Probabilistic Population Codes}
We now consider a population of $n$ neurons, all with (almost) the same tuning curve parameters. The only difference is in the preferred values $p_j$, which are equidistantly spread across the range of possible stimuli (estimation scale).
All parameters determining the population size $n$, the shape of all tuning curves (i.e. $g,w,o$) as well as the presented stimulus $s$ are summarised in a vector $\xi = (n,g,w,o,s)$ which we will refer to as the cognition vector in the following.

Given a particular fixed stimulus $s$, each neuron of this population will respond according to its specific tuning curve and interference due to neural noise. Therefore, the response $r_j$ of the $j$-th neuron must be seen as a realisation of the random variable $R_j\sim \operatorname{Poi}(\lambda=f_{p_j}(s) )$. 
In order to keep in mind, that these responses are always dependent on the parameters of the cognition vector, we henceforth use the notation $r_j(\xi)$ as realisation of $R_j(\xi)$.  The population response is defined by the response of each neuron for a given $\xi$. A singe realisation of such a probabilistic population response is noted as $\varrho(\xi) := \left(r_1(\xi) \, , \ldots , \, r_n(\xi) \right)$. 

\begin{figure}[tb]
    \centering
    \includegraphics[width=\linewidth]{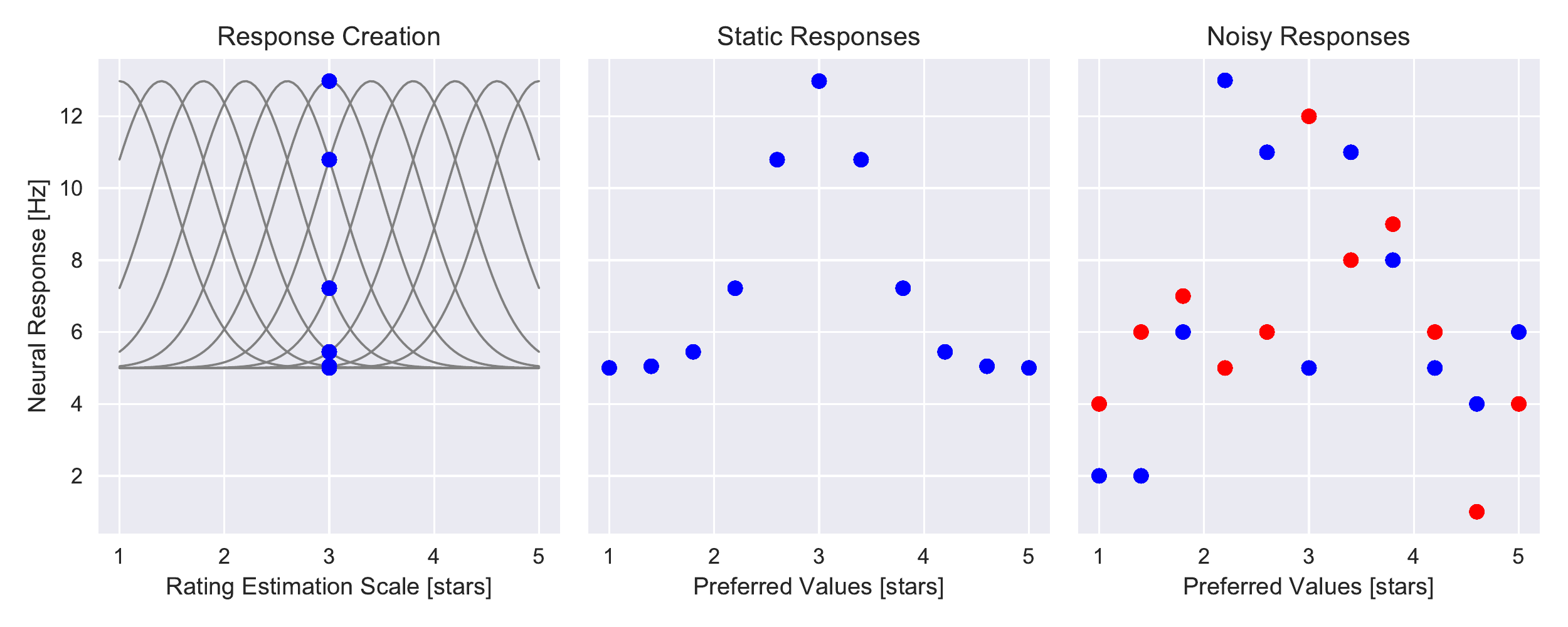}
    \caption{Genesis of noisy population responses demonstrating the alteration for each cognition trial (red and blue).}
    \label{fig:NoisyResponseOrigin}
	\vspace{-2ex}
\end{figure}

This abstract theory of noisy population activity is illustrated in Fig. \ref{fig:NoisyResponseOrigin}.
In this example $n=11$ neurons responded to the stimulus of $s=3$ stars (user rating on a 5-star scale as commonly used by Amazon) where each tuning curve has the offset $o=5\,\text{Hz}$, the width $w=1\,\text{Hz}$ and the gain $g=7$. In the left picture we can see the individual tuning curves, which are distributed equidistantly over the possible range of a five-star rating scale. For $s=3$ stars, the responses of each neuron can be read from its tuning curve. 
For a better representation it has become a standard to plot the individual responses against the corresponding preferred values, which can be seen in the middle picture. These are the theoretical (static) responses without noise. To add neuronal noise, each static response $r^\text{static}_j(\xi)$ is replaced a random number from the Poisson distribution with parameter $\lambda = r^\text{static}_j(\xi)$.
This can be seen in the right subfigure. We additionally repeated this sampling once, i.e. the blue and red dots in each case represent a noisy population response for the same stimulus and it is obvious that these population responses differ not only from the theoretical reference but also very much from each other, i.e. the same cognition leads to different activities on each repetition and each estimate of a choice to be made (e.g. product rating) is thereby given a natural uncertainty.

\subsection{Decoder Functions}
The main question that arises is: How does the brain translate population activity into estimations for a state of the world or a cognition respectively. Theories around the PPC approach assume the use of so-called decoder functions.
Mathematically, a decoder function is a mapping $\varphi\colon\mathbb{R}^n\to S$ from population activity onto the estimation scale for a stimulus or cognition. We will give a brief overview of the most frequently discussed decoder functions in neuroscience literature (exemplified in the context of user feedback):

\begin{figure}[tb]
    \centering
    \includegraphics[width=\linewidth]{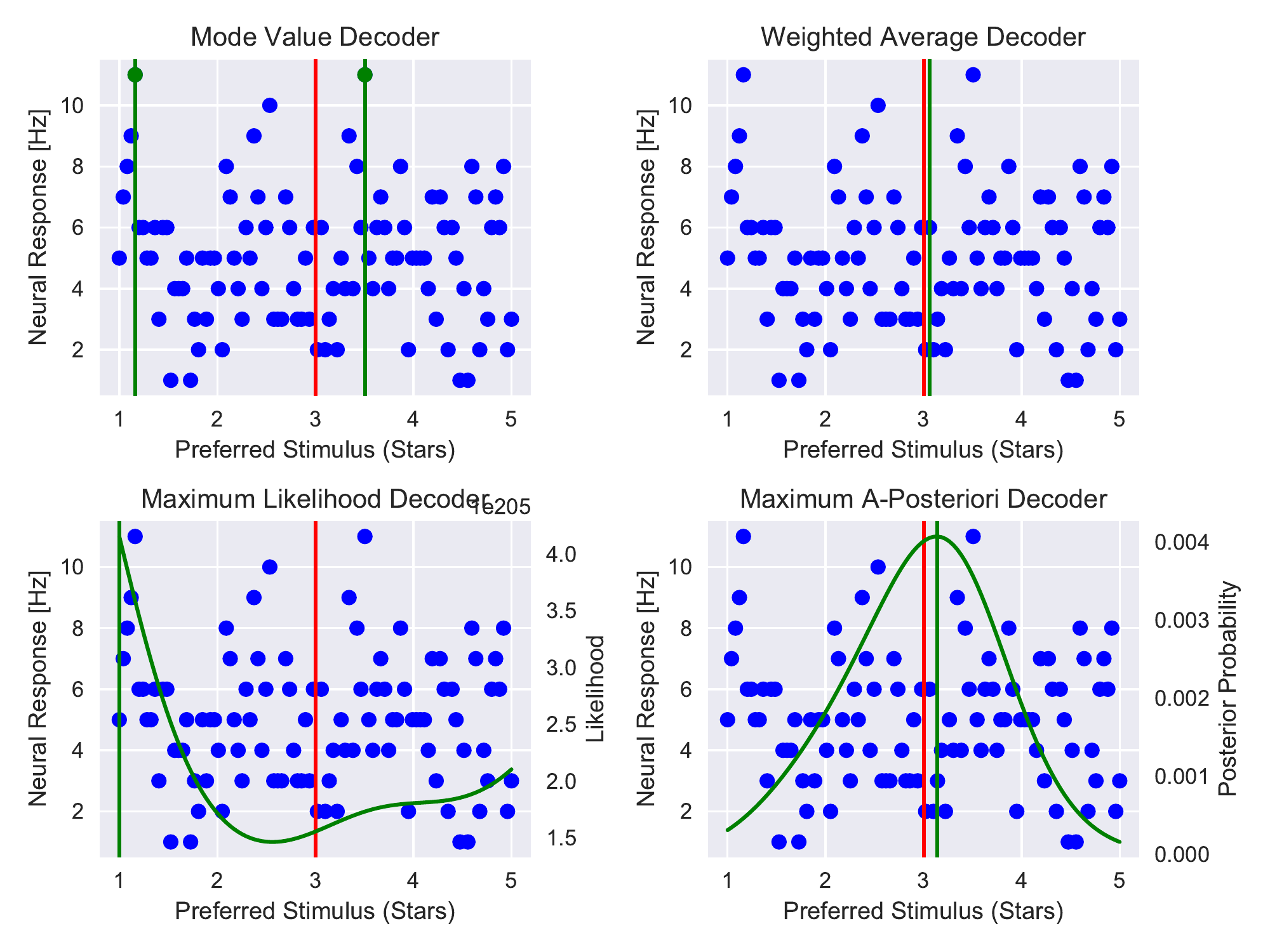}
    \caption{Visualisation of decoder functions on a population response for $\xi=(100,1,1,5,3)$. 
    The red and green lines represent the true and the decoded stimulus. The green graphs show the Likelihood function and the posterior
    density.}
    	\vspace{-2ex}
    \label{fig:FourDecoders}
\end{figure}
\begin{figure}[b]
    \centering
    \includegraphics[width=\linewidth]{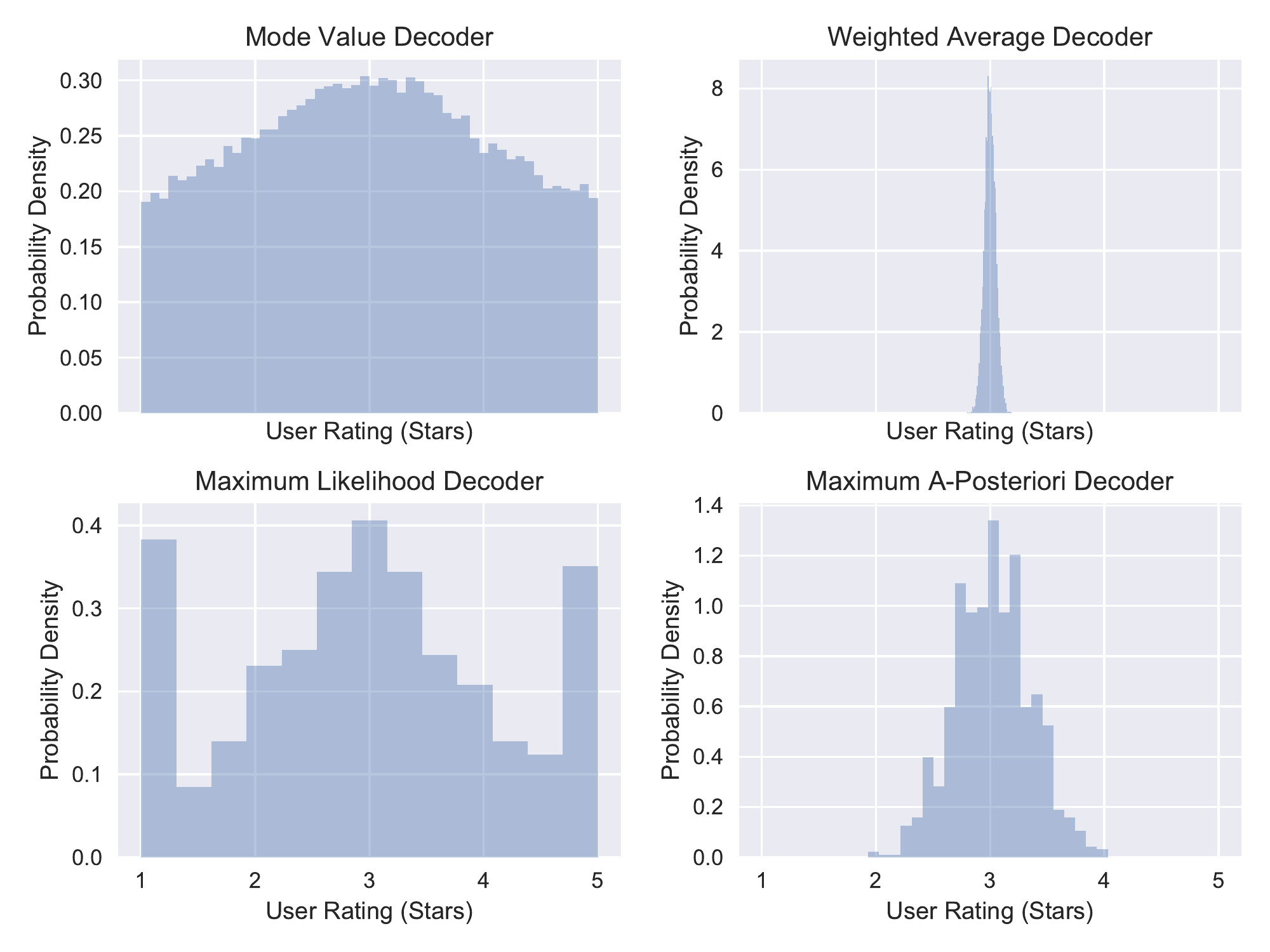}
    \caption{Feedback distributions obtained from different decoder functions for the cognition vector
     $\xi=(100,1,1,5,3)$.}
    \label{fig:FourDists}
	\vspace*{-5ex}
\end{figure}

\subsubsection{Mode Value Decoder (MVD):}
Due to the construction of tuning curves, the MVD assumes that it is exactly the neuron with maximum spiking frequency that is most likely to be addressed by the stimulus or the state of the world. 
Figure \ref{fig:FourDecoders} depicts a population response for a 3-star-decision (red line) together with possible estimators (green lines) for this decision. This decoder is very prone to neuronal noise and its estimators are subject to a great ambiguity which diminishes for higher frequencies in neural responses. The susceptibility to noise is also visible when a stimulus is inferred several times. The resulting feedback distributions for a product rating receives a large variance, as is evident in Fig. \ref{fig:FourDists}.

\vspace{.75ex}

\subsubsection{Weighted Average Decoder (WAD):} The WAD accounts for all responses by setting the specific frequency as a weight to the corresponding preferred value  and considers its contribution to the total response, i.e.
\begin{equation}
\varrho(\xi) \mapsto 
\frac{\sum_{j=1}^n r_j(\xi)\cdot p_j}{\sum_{j=1}^n r_j(\xi)}.
\end{equation}
As to see in Fig. \ref{fig:FourDecoders}, this decoder function provides no ambiguity and is very stable against neuronal noise.

\vspace{.75ex}

\subsubsection{Maximum Likelihood Decoder (MLD):}
For a given population response, the MLD chooses an estimator $s$ with the view to maximise the corresponding likelihood function 
\begin{equation}
P(\varrho(\xi)|s) = \prod_{j=1}^n \frac{f_{p_j}(s)^{r_j(\xi)}}{r_j(\xi)!} 
         \exp\left(-f_{p_j}(s)\right).
\end{equation}
Generally spoken, with the knowledge about the population's tuning curves, a probability density can be computed over the range of possible stimuli which most likely caused a given noisy activity. 
In Fig. \ref{fig:FourDecoders} we see the likelihood function (green curve) for a particular population response together with the maximum likelihood estimator (green line). For lower frequencies, the likelihood often reaches its maximum near the boundaries of a given estimation scale. This heuristically coincides with the observation that users often tend to the extremes of a possible range of choices (e.g. black-and-white mindset). The MLD is the first decoder that explicitly accounts for neuronal noise by processing the Poisson-like distortion of the true tuning curves.

\vspace{.75ex}

\subsubsection{Maximum A Posteriori Decoder (MAD):}
The data-based likelihood can be weighted with prior beliefs $P(s)$ about the stimulus or the states of world which has been learned through former experiences. This is done via Bayes' theorem $P(s|\varrho(\xi)) \propto P(\varrho(\xi)|\hat{s})\cdot P(s)$. 
The estimator is then chosen so that this weighted likelihood (posterior) is maximised.
In Fig. \ref{fig:FourDecoders}, the MAD is much like the MLD but with less variability since the prior knowledge works as a stabiliser. 
Here, we used a Gaussian with $\mu=3$ and $\sigma^2=0.75$ as prior belief, however other priors might be appropriate as well. 
The reliability of inferring a product rating is depicted in Fig. \ref{fig:FourDists} where the resulting distribution does not cover the entire rating scale anymore. It thus describes a restriction of possible choices to be made whilst leaving enough variability to be appropriate for human decision-making at the same time.
The Bayesian brain theory assumes a prominent role of this decoder, since each population would then naturally represent a probability density over a stimulus or state of the world. Further advantages are the simple modelling of learning by using different context-dependent sets of prior beliefs together with a simplified cue integration by Bayesian addition.

\section{Experiments and Evaluation}

In the following, we describe the measurement of real behavioural variability in the context of repeated product ratings.
We then find a corresponding cognition vector by using stochastic simulations and information-theoretical metrics.
In other words, we draw conclusions from the observed distribution to the underlying states of the population tuning curves.
We then examine to what extent these conclusions are consistent with previous neurological findings. For the purpose of reproducibility and to support further research, explicit algorithms and datasets are publicly available for download at: www.double-blind.edu.

\subsection{Measuring Decision Variability (User Study)}
\begin{figure}[tb]
    \centering
    \begin{subfigure}{0.49\linewidth}
        \includegraphics[width=\textwidth]{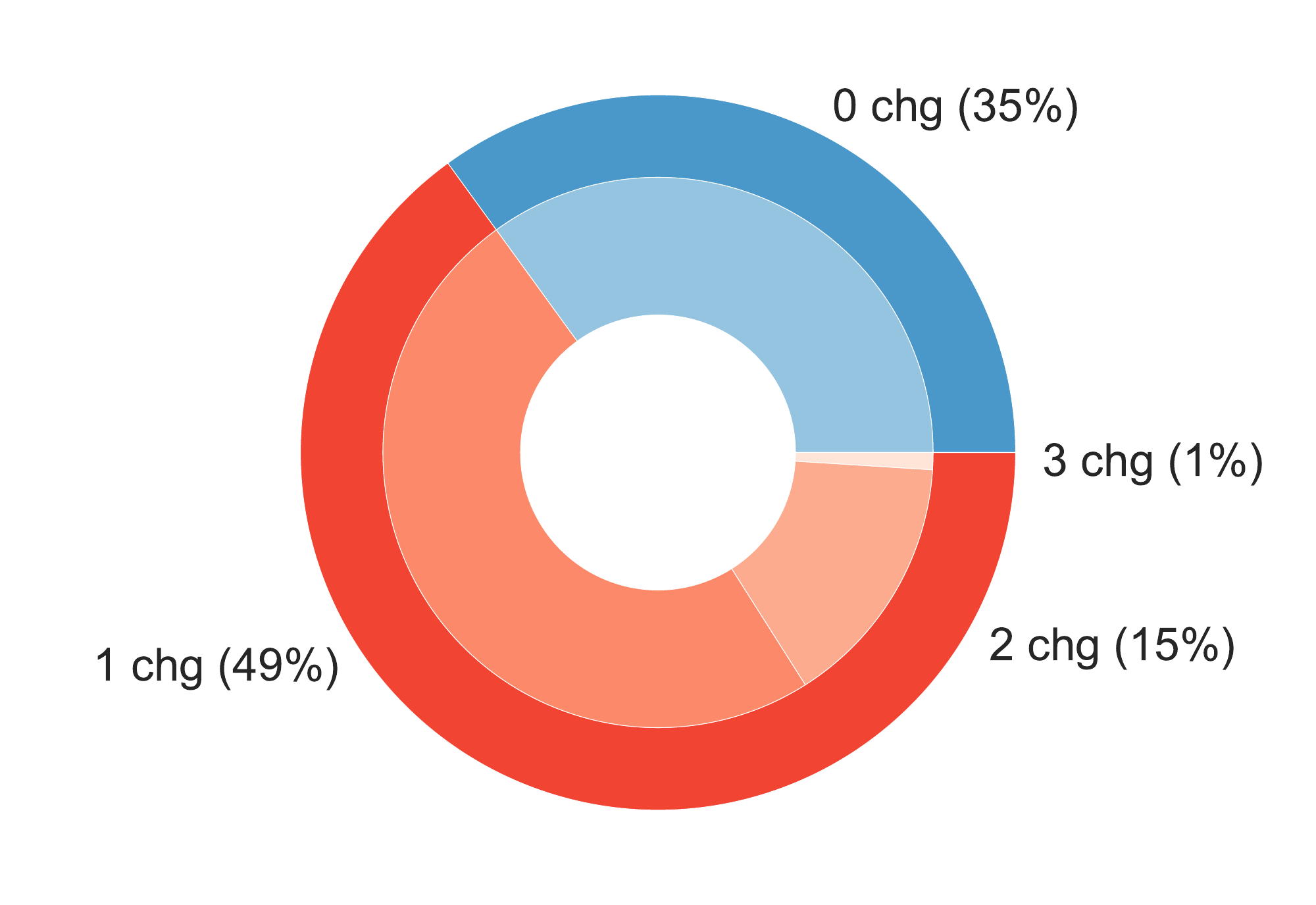}
        \caption{frequency of choice changes}
        \label{fig:ResponseCats}
    \end{subfigure}
    \hfill
    \begin{subfigure}{0.49\linewidth}
        \includegraphics[width=\textwidth]{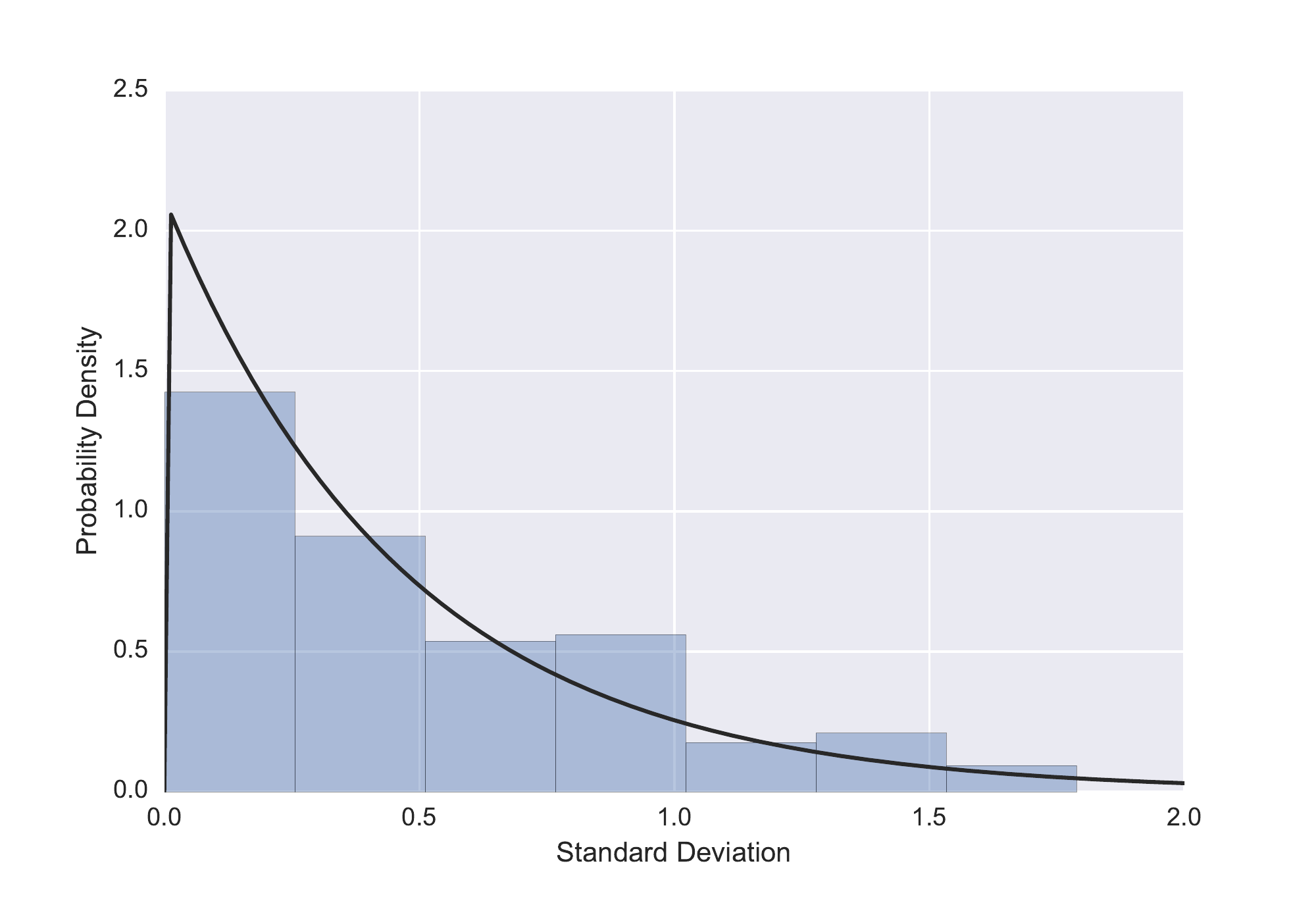}
        \caption{distribution of variances}
         \label{fig:DistVar}
    \end{subfigure}
    \vspace{-1ex}
    \caption{Visualisation of behavioural variability and decision changes found in the RETRAIN user study.}
    \vspace{-2ex}
\end{figure}

We conducted the RETRAIN (Reliability Trailer Rating) study as an online experiment in which 67 participants had watched theatrical trailers of popular movies and television shows and provided ratings in five consecutive repetition trials. User ratings have been recorded for five of ten trailers so that the remaining ones acted as distractors, triggering the misinformation effect, i.e. memory is becoming less accurate due to interference from post-event information. The so obtained data set comprises $N = 1\,675$ individual ratings. 
User responses scattered around a central tendency rather than being constant, i.e. from all user ratings, only 35\% manifested a consistent response behaviour, while 50\% gave two different responses on the same item, and 15\% used even three or more different ratings (see Fig. \ref{fig:ResponseCats}). The extent of variability (standard deviations) is exponentially distributed as to see in Fig. \ref{fig:DistVar}.
In the following, we use this data record to fit individual feedback distributions. These will then be compared with our model-based distributions generated by the PPC approach.

\subsection{Fitting Cognition Vectors}
After measuring real human decisions and its variability we have to assign each of these to a theoretical correspondence in the form of a cognition vector. 
For this purpose, we restrict each parameter to the intervals $n\in [25,250]$, $g\in [1,100]$, $w\in [0.1,2.0]$, $o\in [1,15]$, $s\in [1,5]$
and subdivide these into ten equidistant values. This gives us $10^5$ different combinations which, in conjunction with each of the four decoder functions, are converted into a feedback distribution brute force. 
The conversion is done analogously to generating the distributions of Fig. \ref{fig:FourDists} by repeating the decoding of noisy population activities.
To find the model-based distribution $P_{\text{model}}$ that is closest to the real distribution $P_{\text{real}}$, we use a similarity metric of information theory, the Jensen-Shannon divergence $0 \leq \operatorname{JSD}(P_{\text{model}},P_{\text{real}}) \leq 1$
where $0$ stands for perfect agreement \cite{jsd}. With this method we are able to identify the cognition vector that best reproduces reality.

\subsection{Analysing Adequacy}
In order for the PPC to be recognised as an adequate model for human cognition and decision-making, it has to match the measurement results of medical correspondences from other publications in the field of neuroscience. Therefore, we evaluate the following properties for each best-fit cognition vector and compare these to the results of other publications or common knowledge in the respective fields of research:
\begin{enumerate}
\item  It is an axiomatic imperative that the model is able to generate the real probability densities at all. For this reason, we do not only select the cognition vectors with the lowest JSD score, but consider the distribution of best-fit scores. Furthermore, the resulting distributions are classified by observers as appropriate or inappropriate.

\item The cognition vectors are retroactively converted into neuronal activity (spiking rates) to check whether they are within the biological limitations of neuronal cells. Combined effects of these frequencies are crucial to populations. So we simulate the events over time for each cognition vector and assess the constructive interference. From this, possible EEG waves can be deduced.

\item  The interaction of tuning curve parameters should not be arbitrary but condition each other. Therefore, the correlations of the respective parameters for the best-fit cognition vectors are examined.
\end{enumerate}

\section{Results}
\begin{figure}[t]
    \centering
    \includegraphics[width=.9\linewidth]{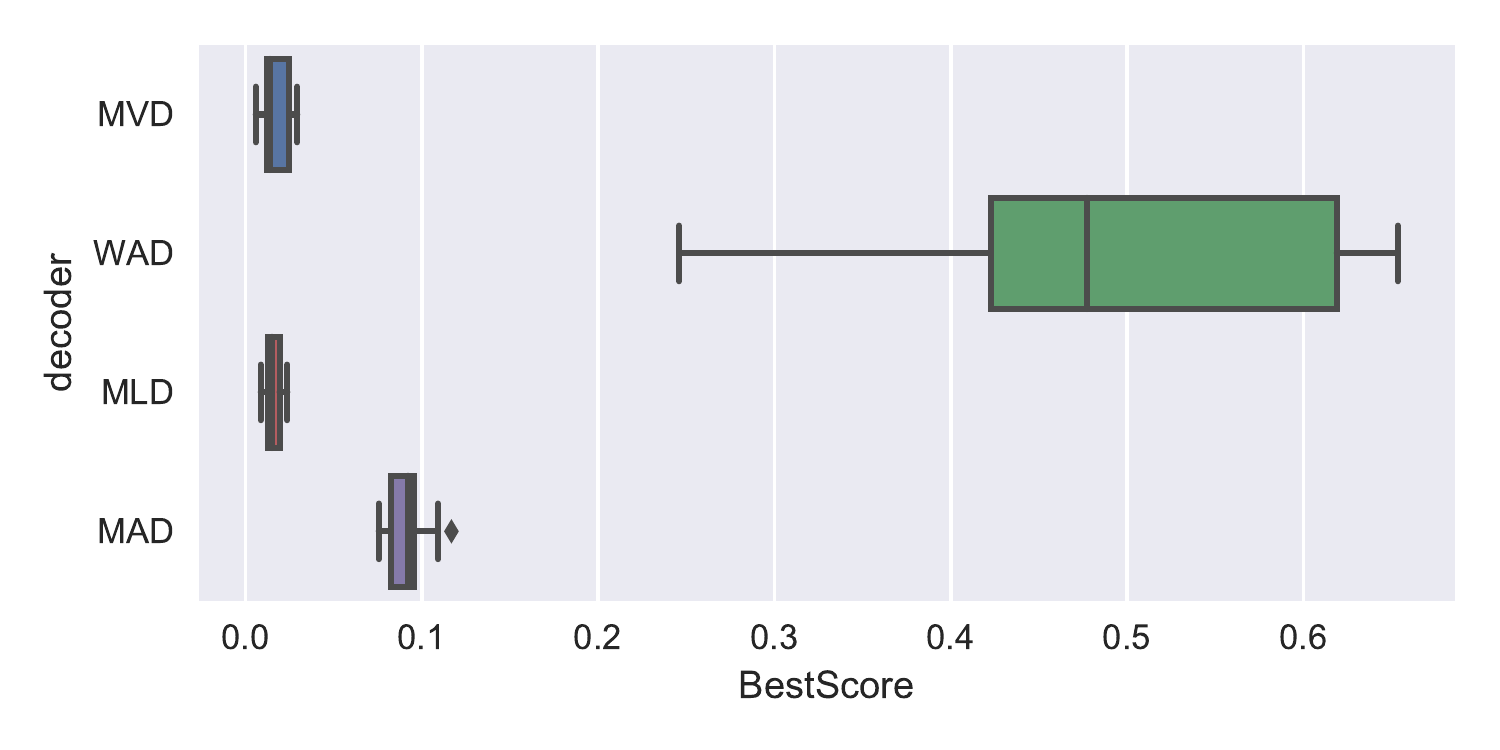}
    \caption{Best similarity scores for model-based and real user feedback. The smaller the score, the better the adaptation.}
    \label{fig:BestScores}
	\vspace{-2ex}
\end{figure}

A comparison of the best-fit JSD scores reveals that the MVD and MLD decoder are good choices of modelling user behaviour (see Fig. \ref{fig:BestScores}). As already seen in Fig. \ref{fig:FourDists}, the WAD decoder is too precise to cover behavioural variability. However, it is surprising, that the MAD along with the real user feedback as prior knowledge performs worse than the MLD. This might be a hint that no prior knowledge was used for decision-making since the MLD is a special case of the MAD with an uninformative prior.
For each decoder function, we plotted all the model-based feedback together with the real distributions.
Three independent experts rated the adjustments as appropriate or inappropriate.
The MLD indeed performed best (95\% appropriates, 92\% inter-rater agreement) over the MVD (78\% appropriates, 89\% IRA) as well as the MAD (13\% appropriates, 98\% IRA) and the WAD (2\% appropriates, 99\% IRA).

\begin{figure}[tb]
    \centering
    \begin{subfigure}{0.49\linewidth}
        \includegraphics[width=\textwidth]{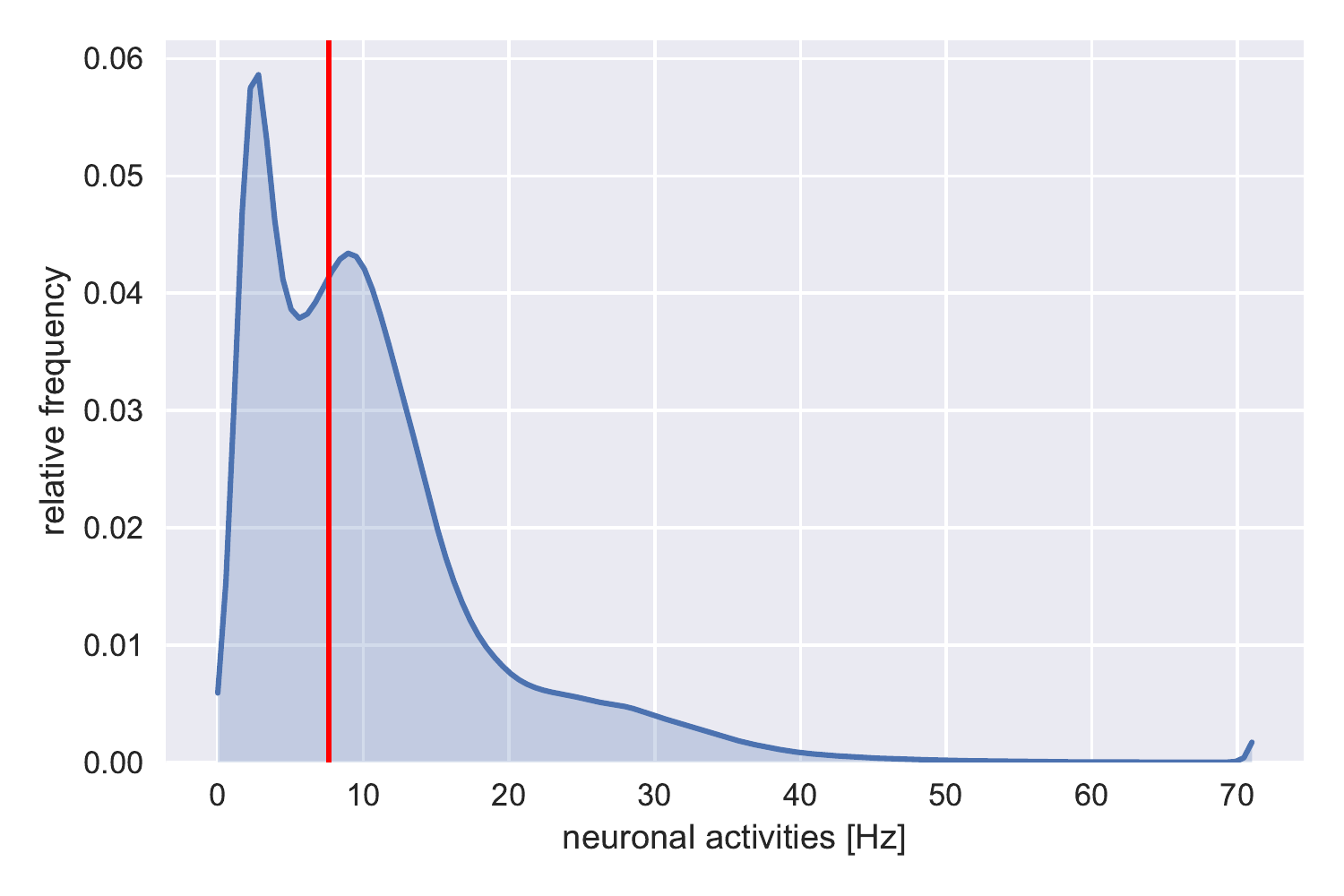}
        \caption{neuronal activity distribution}
        \label{fig:activity}
    \end{subfigure}
    \hfill
    \begin{subfigure}{0.49\linewidth}
        \includegraphics[width=\textwidth]{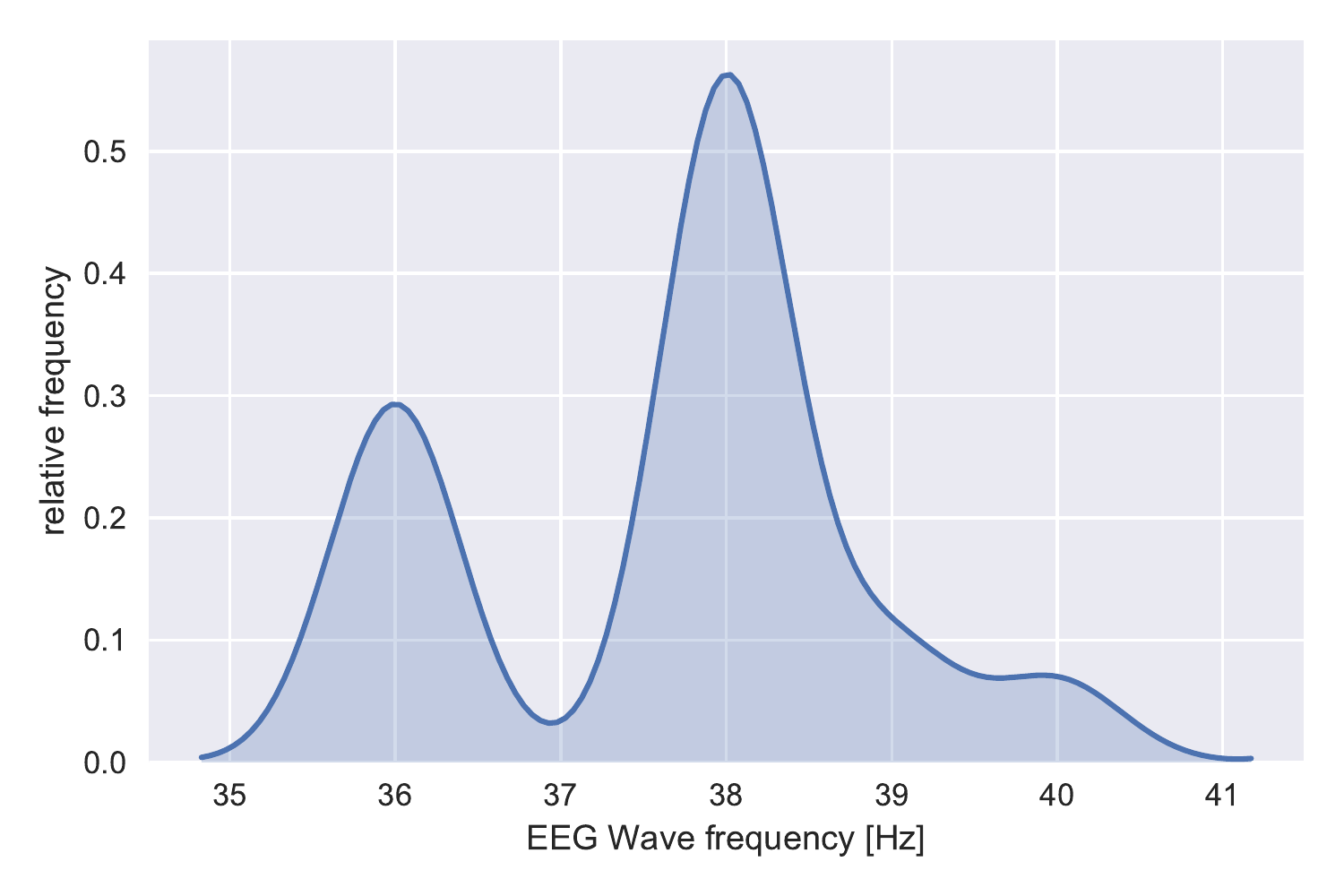}
        \caption{EEG frequency distribution}
         \label{fig:eeg}
    \end{subfigure}
    \vspace{-1ex}
    \caption{Occurrence of frequencies and their interference for simulated decision-making using the MLD.}
    \vspace{-2ex}
\end{figure}

\begin{figure}[b]
    \centering
    \includegraphics[width=\linewidth]{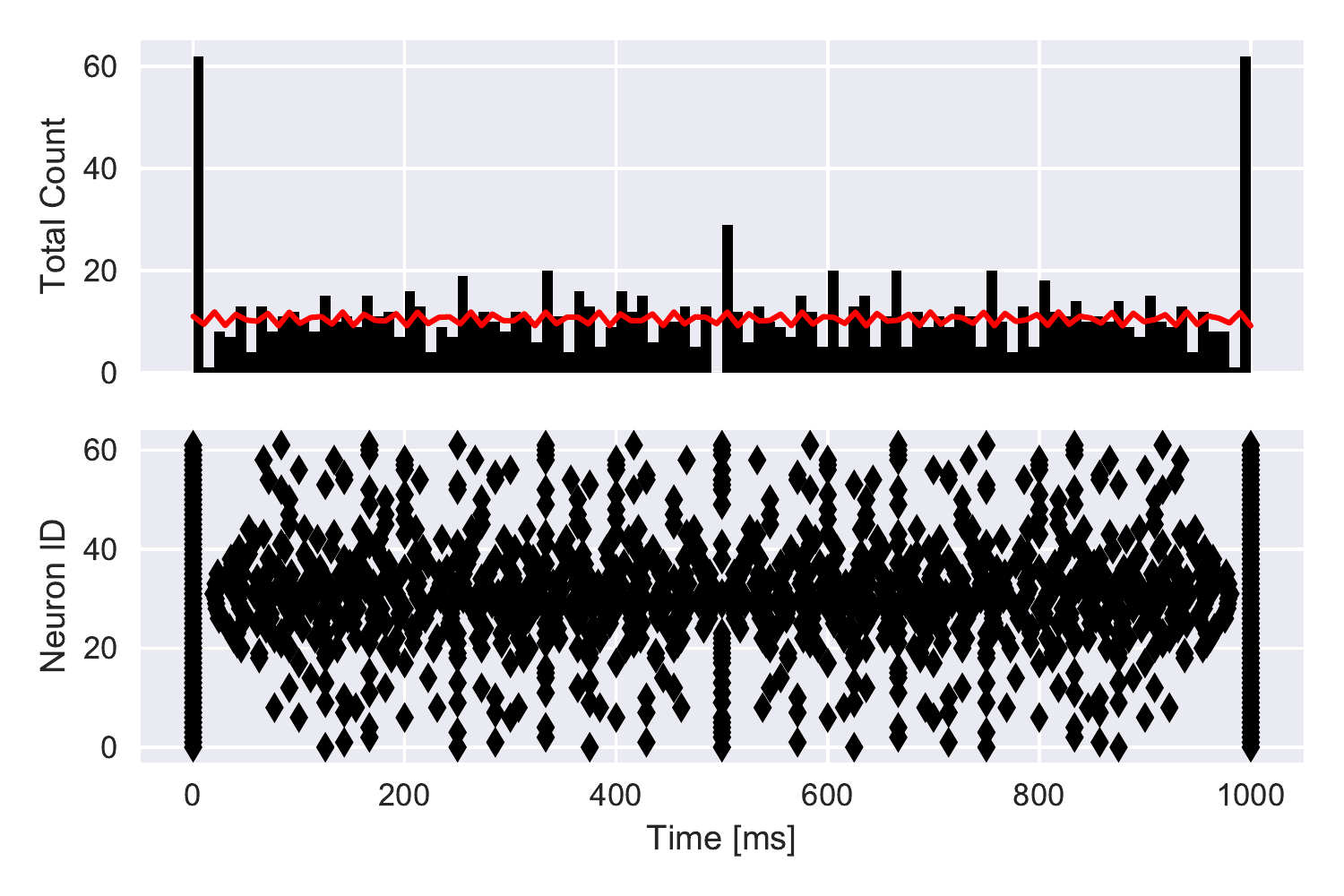}
    \caption{Events over time for a particular cognition vector. Total counts can be fitted using a sine function (red curve) which allows to
     estimate a frequency for possible EEG waves.}
    \label{fig:PAOT}
	\vspace{-2ex}
\end{figure}

For a given cognition vector and the MLD as decoder function, the population spiking frequencies have been re-computed. The distribution of spiking rates for all product ratings is depicted in Fig. \ref{fig:activity}. We observe a log-norm distribution between 0 and 70 Hz and expectation of about 8 Hz (red line). Medical contributions also often report a log-norm distribution of neural activities, which usually ranges between 0 and 100 Hz with a mean of 10 Hz \cite{freq1}. The configured model of the PPC hence leads to neural activities which are close to those usually reported in real measurements. 

Having specific spiking frequencies, the resulting events over time (exemplified in Fig.\ref{fig:PAOT}) informs about constructive interferences of the individual spikes. The total counts can be fitted using a sine function and due to the proportionality of simultaneous spikes to the induced electrical current, this sine function can be used to estimate frequencies for corresponding EEG waves. The distribution of those EEG frequencies for all best-fit cognition vectors using the MLD is depicted in Fig. \ref{fig:eeg}.
We recognise that all cognitions took place within the lower gamma band (30-100 Hz, but usually not exceeding 40 Hz for normal conditions). The gamma band is associated to demanding activities with a high flow of information and are additionally expected during cross-modal sensory processing (e.g. combining sound and sight) \cite{66}. The participants involved have been watching video trailers (i.e. sound and sight) and formed an opinion at the same time while trying to remember previous answers. EEG waves within the gamma band are thus in perfect accordance with common expectations. 
\pagebreak

\begin{figure}[tb]
    \centering
    \includegraphics[width=\linewidth]{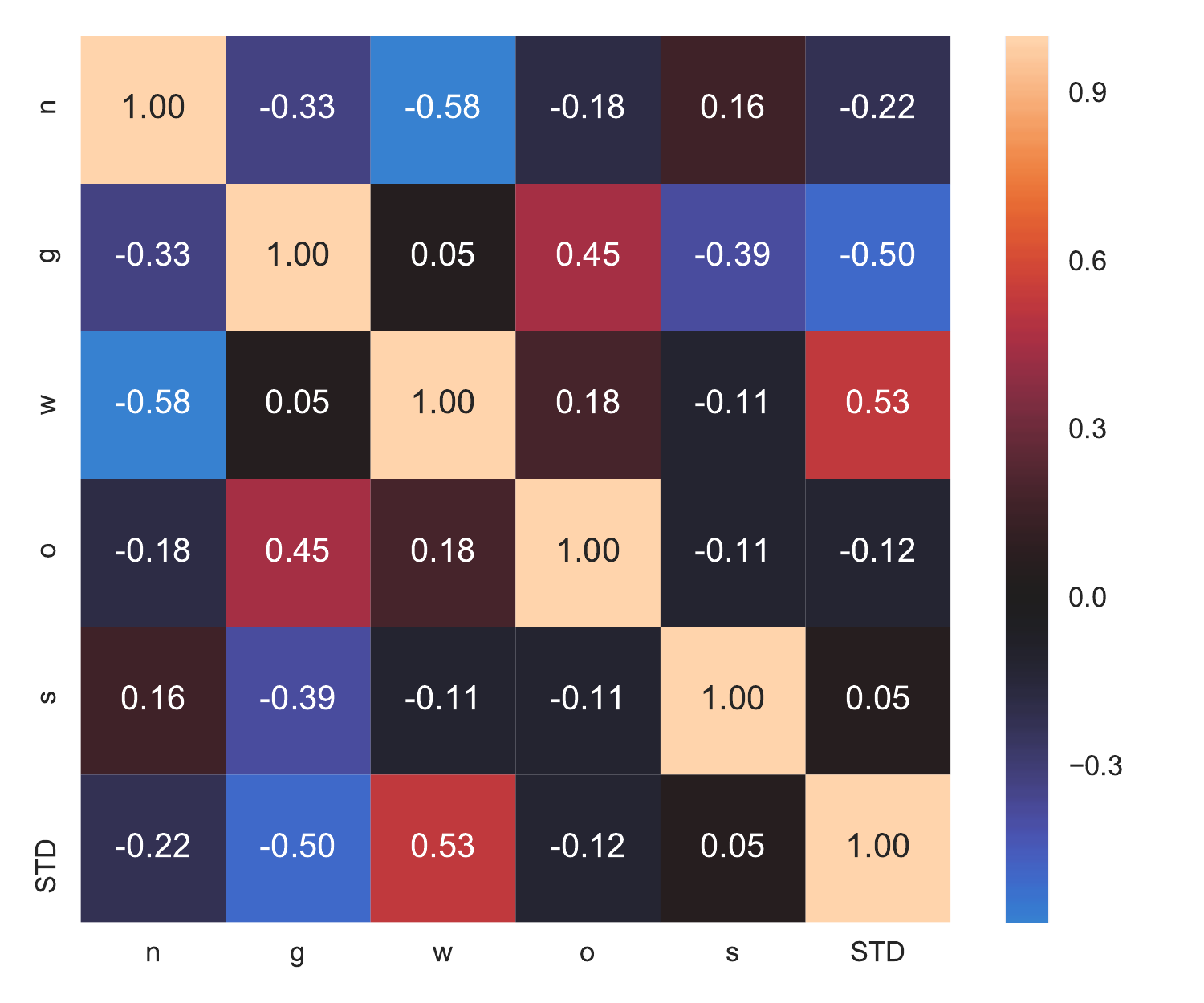}
    \vspace*{-5ex}
    \caption{Correlation heatmap of tuning curve parameters for best-fit cognition vectors using the MLD.}
    \label{fig:Corrs}
	\vspace{-2ex}
\end{figure}

The interplay of tuning curve parameters and the uncertainty of user feedback (i.e. standard deviation STD) is depicted by the correlation heatmap in Fig. \ref{fig:Corrs}. The behavioural variability (STD) mainly correlates with the frequency gain $g$ and the tuning curve width $w$. This is not surprising, because we utilise a Gaussian density for the tuning curve shape and this automatically outputs a larger response frequency for smaller widths due to the normalisation. To be able to further adjust this frequency despite a fixed width, we introduced the gain as an additional stretching factor. In principle, the following is apparent: the greater the frequency, the less impact is given the neuronal noise and the smaller the standard deviation in the repetition of decision-making.
There are further correlative pairs given by $n$-$w$ and $g$-$o$.
The negative correlation between $n$ and $w$ is only logical, because if more tuning curves are distributed equidistantly on a fixed scale, then their widths no longer have to be that large  to completely cover this scale. Here, the model automatically proceeded according to logical schemes. The positive correlation between $g$ and $o$ is a non-obvious property of the model. If higher activities are needed, this is not done only by raising the gain (fast reduction of behavioural variability) but also by raising the offset (ground spiking rate).

\section{Discussion}
In this contribution, we have merged the model of PPC with real-world data from predictive data mining for the variability of decision-making. Simulations have shown that this hitherto purely theoretical model can be configured in such a way that it is able to explain the real behaviour of study participants almost perfectly. This configured model then made conclusions about spiking activity and EEG waves, which almost exactly meet the expectations deduced from real measurements in the field of neuroscience. However, these medical correlates have never been directly entered into this PPC model but emerged naturally by simply tailoring the model-based feedback to the real user feedback from a repeated trailer rating scenario. In addition, the model has plausible correlations, as they are assumed in this context.

These findings provide evidence that the model of PPC is not only suitable for sensory perception and motor control, but can also explain higher cognitions and decision-making along with its immanent variability. The involvement of PPC in such tasks of cognition has always been suspected but has never been explicitly studied so far. To this end, our contribution introduces a methodology that is capable of accessing these research questions for further investigations.

\section{Further Research}
During our analysis, we also tried sigmoid-shaped tuning curves and found that the population activity forms monotonically increasing and decreasing curves, even with asymptotes depending on the particular configuration. These curves are often used in utility theory. Since PPCs come into question as a model of decision-making, the additional ability for describing utility functions would be a lucky ``coincidence''. This is utterly important to investigate through further research.

%\section{Acknowledgments}
%Computational support and infrastructure was provided by the “Centre for Information and Media Technology” (ZIM) at the University of Düsseldorf (Germany). 

\bibliographystyle{apacite}
\setlength{\bibleftmargin}{.125in}
\setlength{\bibindent}{-\bibleftmargin}
\bibliography{CogSci_Template}

\begin{thebibliography}{}

\bibitem [\protect \citeauthoryear {%
Chapin%
}{%
Chapin%
}{%
{\protect \APACyear {2004}}%
}]{%
motor}
\APACinsertmetastar {%
motor}%
\begin{APACrefauthors}%
Chapin, J\BPBI K.%
\end{APACrefauthors}%
\unskip\
\newblock
\APACrefYearMonthDay{2004}{}{}.
\newblock
{\BBOQ}\APACrefatitle {Using multi-neuron population recordings for neural
  prosthetics} {Using multi-neuron population recordings for neural
  prosthetics}.{\BBCQ}
\newblock
\APACjournalVolNumPages{Nature neuroscience}{7}{5}{}.
\PrintBackRefs{\CurrentBib}

\bibitem [\protect \citeauthoryear {%
Dayan%
\ \BBA {} Abbott%
}{%
Dayan%
\ \BBA {} Abbott%
}{%
{\protect \APACyear {2001}}%
}]{%
affe}
\APACinsertmetastar {%
affe}%
\begin{APACrefauthors}%
Dayan, P.%
\BCBT {}\ \BBA {} Abbott, L.%
\end{APACrefauthors}%
\unskip\
\newblock
\APACrefYear{2001}.
\newblock
\APACrefbtitle {Theoretical neuroscience} {Theoretical neuroscience}.
\newblock
\APACaddressPublisher{}{Cambridge, MA: MIT Press}.
\PrintBackRefs{\CurrentBib}

\bibitem [\protect \citeauthoryear {%
Faisal%
, Selen%
\BCBL {}\ \BBA {} Wolpert%
}{%
Faisal%
\ \protect \BOthers {.}}{%
{\protect \APACyear {2008}}%
}]{%
nervousnoise}
\APACinsertmetastar {%
nervousnoise}%
\begin{APACrefauthors}%
Faisal, A\BPBI A.%
, Selen, L\BPBI P.%
\BCBL {}\ \BBA {} Wolpert, D\BPBI M.%
\end{APACrefauthors}%
\unskip\
\newblock
\APACrefYearMonthDay{2008}{}{}.
\newblock
{\BBOQ}\APACrefatitle {Noise in the nervous system} {Noise in the nervous
  system}.{\BBCQ}
\newblock
\APACjournalVolNumPages{Nature reviews neuroscience}{9}{4}{}.
\PrintBackRefs{\CurrentBib}

\bibitem [\protect \citeauthoryear {%
Hill%
}{%
Hill%
}{%
{\protect \APACyear {1995}}%
}]{%
Hill}
\APACinsertmetastar {%
Hill}%
\begin{APACrefauthors}%
Hill, W.%
\end{APACrefauthors}%
\unskip\
\newblock
\APACrefYearMonthDay{1995}{}{}.
\newblock
{\BBOQ}\APACrefatitle {Recommending and Evaluating Choices} {Recommending and
  evaluating choices}.{\BBCQ}
\newblock
\BIn{} \APACrefbtitle {SIGCHI Conference.} {Sigchi conference.}
\PrintBackRefs{\CurrentBib}

\bibitem [\protect \citeauthoryear {%
Jannach%
\ \BBA {} Zanker%
}{%
Jannach%
\ \BBA {} Zanker%
}{%
{\protect \APACyear {2010}}%
}]{%
Jannach}
\APACinsertmetastar {%
Jannach}%
\begin{APACrefauthors}%
Jannach, D.%
\BCBT {}\ \BBA {} Zanker, M.%
\end{APACrefauthors}%
\unskip\
\newblock
\APACrefYear{2010}.
\newblock
\APACrefbtitle {Recommender Systems: An Introduction} {Recommender systems: An
  introduction}.
\newblock
\APACaddressPublisher{}{Cambridge University Press}.
\PrintBackRefs{\CurrentBib}

\bibitem [\protect \citeauthoryear {%
Jasberg%
\ \BBA {} Sizov%
}{%
Jasberg%
\ \BBA {} Sizov%
}{%
{\protect \APACyear {2017}}%
}]{%
JasUMAP}
\APACinsertmetastar {%
JasUMAP}%
\begin{APACrefauthors}%
Jasberg, K.%
\BCBT {}\ \BBA {} Sizov, S.%
\end{APACrefauthors}%
\unskip\
\newblock
\APACrefYearMonthDay{2017}{}{}.
\newblock
{\BBOQ}\APACrefatitle {Probabilistic Perspectives on Collecting Human
  Uncertainty in Predictive Data Mining} {Probabilistic perspectives on
  collecting human uncertainty in predictive data mining}.{\BBCQ}
\newblock
\APACjournalVolNumPages{UMAP Conference}{}{}{}.
\PrintBackRefs{\CurrentBib}

\bibitem [\protect \citeauthoryear {%
Kanayama%
, Sato%
\BCBL {}\ \BBA {} Ohira%
}{%
Kanayama%
\ \protect \BOthers {.}}{%
{\protect \APACyear {2007}}%
}]{%
66}
\APACinsertmetastar {%
66}%
\begin{APACrefauthors}%
Kanayama, N.%
, Sato, A.%
\BCBL {}\ \BBA {} Ohira, H.%
\end{APACrefauthors}%
\unskip\
\newblock
\APACrefYearMonthDay{2007}{}{}.
\newblock
{\BBOQ}\APACrefatitle {Crossmodal effect with rubber hand illusion and
  gamma-band activity} {Crossmodal effect with rubber hand illusion and
  gamma-band activity}.{\BBCQ}
\newblock
\APACjournalVolNumPages{Psychophysiology}{44}{3}{pp. 392--402}.
\PrintBackRefs{\CurrentBib}

\bibitem [\protect \citeauthoryear {%
Knill%
\ \BBA {} Pouget%
}{%
Knill%
\ \BBA {} Pouget%
}{%
{\protect \APACyear {2004}}%
}]{%
cueintegration}
\APACinsertmetastar {%
cueintegration}%
\begin{APACrefauthors}%
Knill, D\BPBI C.%
\BCBT {}\ \BBA {} Pouget, A.%
\end{APACrefauthors}%
\unskip\
\newblock
\APACrefYearMonthDay{2004}{}{}.
\newblock
{\BBOQ}\APACrefatitle {The Bayesian brain: the role of uncertainty in neural
  coding and computation} {The bayesian brain: the role of uncertainty in
  neural coding and computation}.{\BBCQ}
\newblock
\APACjournalVolNumPages{TRENDS in Neurosciences}{27}{12}{pp. 712--719}.
\PrintBackRefs{\CurrentBib}

\bibitem [\protect \citeauthoryear {%
Lin%
}{%
Lin%
}{%
{\protect \APACyear {1991}}%
}]{%
jsd}
\APACinsertmetastar {%
jsd}%
\begin{APACrefauthors}%
Lin, J.%
\end{APACrefauthors}%
\unskip\
\newblock
\APACrefYearMonthDay{1991}{}{}.
\newblock
{\BBOQ}\APACrefatitle {Divergence measures based on the Shannon entropy}
  {Divergence measures based on the shannon entropy}.{\BBCQ}
\newblock
\APACjournalVolNumPages{IEEE Transactions on Information theory}{37}{1}{}.
\PrintBackRefs{\CurrentBib}

\bibitem [\protect \citeauthoryear {%
Ma%
}{%
Ma%
}{%
{\protect \APACyear {2009}}%
}]{%
decoders}
\APACinsertmetastar {%
decoders}%
\begin{APACrefauthors}%
Ma, W.%
\end{APACrefauthors}%
\unskip\
\newblock
\APACrefYearMonthDay{2009}{}{}.
\newblock
{\BBOQ}\APACrefatitle {Population Codes: theoretic aspects} {Population codes:
  theoretic aspects}.{\BBCQ}
\newblock
\APACjournalVolNumPages{Encyclopedia of neuroscience}{}{}{}.
\PrintBackRefs{\CurrentBib}

\bibitem [\protect \citeauthoryear {%
Moser%
, Kropff%
\BCBL {}\ \BBA {} Moser%
}{%
Moser%
\ \protect \BOthers {.}}{%
{\protect \APACyear {2008}}%
}]{%
placecells}
\APACinsertmetastar {%
placecells}%
\begin{APACrefauthors}%
Moser, E\BPBI I.%
, Kropff, E.%
\BCBL {}\ \BBA {} Moser, M\BHBI B.%
\end{APACrefauthors}%
\unskip\
\newblock
\APACrefYearMonthDay{2008}{}{}.
\newblock
{\BBOQ}\APACrefatitle {Place cells, grid cells, and the brain's spatial
  representation system} {Place cells, grid cells, and the brain's spatial
  representation system}.{\BBCQ}
\newblock
\APACjournalVolNumPages{Annu. Rev. Neurosci.}{31}{}{pp. 69--89}.
\PrintBackRefs{\CurrentBib}

\bibitem [\protect \citeauthoryear {%
Pouget%
}{%
Pouget%
}{%
{\protect \APACyear {2006}}%
}]{%
Pouget}
\APACinsertmetastar {%
Pouget}%
\begin{APACrefauthors}%
Pouget, A.%
\end{APACrefauthors}%
\unskip\
\newblock
\APACrefYearMonthDay{2006}{}{}.
\newblock
{\BBOQ}\APACrefatitle {Bayesian inference with probabilistic population codes}
  {Bayesian inference with probabilistic population codes}.{\BBCQ}
\newblock
\APACjournalVolNumPages{Nature Neuroscience}{}{9}{}.
\PrintBackRefs{\CurrentBib}

\bibitem [\protect \citeauthoryear {%
Ricci%
}{%
Ricci%
}{%
{\protect \APACyear {2015}}%
}]{%
Handbook}
\APACinsertmetastar {%
Handbook}%
\begin{APACrefauthors}%
Ricci, F.%
\end{APACrefauthors}%
\unskip\
\newblock
\APACrefYear{2015}.
\newblock
\APACrefbtitle {Recommender Systems Handbook} {Recommender systems handbook}.
\newblock
\APACaddressPublisher{}{Springer}.
\PrintBackRefs{\CurrentBib}

\bibitem [\protect \citeauthoryear {%
Roxin%
, Brunel%
, Hansel%
, Mongillo%
\BCBL {}\ \BBA {} van Vreeswijk%
}{%
Roxin%
\ \protect \BOthers {.}}{%
{\protect \APACyear {2011}}%
}]{%
freq1}
\APACinsertmetastar {%
freq1}%
\begin{APACrefauthors}%
Roxin, A.%
, Brunel, N.%
, Hansel, D.%
, Mongillo, G.%
\BCBL {}\ \BBA {} van Vreeswijk, C.%
\end{APACrefauthors}%
\unskip\
\newblock
\APACrefYearMonthDay{2011}{}{}.
\newblock
{\BBOQ}\APACrefatitle {On the distribution of firing rates in networks of
  cortical neurons} {On the distribution of firing rates in networks of
  cortical neurons}.{\BBCQ}
\newblock
\APACjournalVolNumPages{Journal of Neuroscience}{31}{45}{pp. 16217--16226}.
\PrintBackRefs{\CurrentBib}

\end{thebibliography}

\end{document}